\documentclass[letterpaper, 10 pt, conference]{ieeeconf}  

\IEEEoverridecommandlockouts                              

\usepackage{graphics} 
\usepackage{epsfig} 
\usepackage{mathptmx} 
\usepackage{times} 
\usepackage{amsmath} 
\usepackage{amssymb}  
\usepackage{color}
\usepackage{threeparttable}
\title{\LARGE \bf
Underwater Intention Recognition using Head Motion and \\
Throat Vibration for Supernumerary Robotic Assistance*
}
\author{
    Yuqin Guo$^{1,2,\#}$, 
    Rongzheng Zhang$^{3,\#}$, 
    Wanghongjie Qiu$^{2}$,      
    Harry Asada$^{4}$,~\IEEEmembership{Fellow,~IEEE,}\\
    Fang Wan$^{1,3,*}$,~\IEEEmembership{Member,~IEEE} and 
    Chaoyang Song$^{2,5,*}$,~\IEEEmembership{Senior~Member,~IEEE}
\thanks{
    *This work was partly supported by the SUSTech-MIT Joint Centers for Mechanical Engineering Research and Education, the Science, Technology, and Innovation Commission of Shenzhen Municipality [ZDSYS20220527171403009, JCYJ20220818100417038, 20200925155748006], National Science Foundation of China [62206119], Guangdong Provincial Key Laboratory of Human-Augmentation and Rehabilitation Robotics in Universities. Y. Guo and R. Zhang are co-first authors. F. Wan and C. Song are co-corresponding authors: {\tt\small wanf@sustech.edu.cn}, {\tt\small songcy@ieee.org}
    }
\thanks{$^{1}$Shenzhen Key Laboratory of Intelligent Robotics and Flexible Manufacturing, Southern University of Science and Technology, 
        Shenzhen, Guangdong 518055, China.}%
\thanks{$^{2}$Department of Mechanical and Energy Engineering, Southern University of Science and Technology,
        Shenzhen, Guangdong 518055, China.}%
\thanks{$^{3}$School of Design, Southern University of Science and Technology,
        Shenzhen, Guangdong 518055, China.}%
\thanks{$^{4}$Department of Mechanical Engineering, Massachusetts Institute of Technology,
        Cambridge, MA 02139, USA.}%
\thanks{$^{5}$SUSTech Institute of Robotics, Southern University of Science and Technology,
        Shenzhen, Guangdong 518055, China.}%
}
\begin{document}
\maketitle
\thispagestyle{empty}
\pagestyle{empty}
\begin{abstract}

    This study presents a multi-modal mechanism for recognizing human intentions while diving underwater, aiming to achieve natural human-robot interactions through an underwater superlimb for diving assistance. The underwater environment severely limits the divers' capabilities in intention expression, which becomes more challenging when they intend to operate tools while keeping control of body postures in 3D with the various diving suits and gears. The current literature is limited in underwater intention recognition, impeding the development of intelligent wearable systems for human-robot interactions underwater. Here, we present a novel solution to simultaneously detect head motion and throat vibrations under the water in a compact, wearable design. Experiment results show that using machine learning algorithms, we achieved high performance in integrating these two modalities to translate human intentions to robot control commands for an underwater superlimb system. This study's results paved the way for future development in underwater intention recognition and underwater human-robot interactions with supernumerary support.
    
\end{abstract}

\section{Introduction}
\label{sec:Introduction}

    Diving with Self-Contained Underwater Breathing Apparatus (SCUBA) is a popular activity for exploring the ocean, which involves a series of professional equipment wearable on the human body for life-support and body movement \cite{straughan2012touched}. However, the level of intelligence of these diving gears remains primarily mechanical by design. There remains a research gap in introducing robotic solutions toward autonomous, natural interactions between human divers and the underwater environment, where novel designs in wearable robots and interactive mechanisms need further exploration \cite{xia2022wearable}.

    Before introducing wearable robots to assist human divers, intention recognition underwater becomes a critical issue due to challenges brought by the aquatic environment. Currently, hand gestures are the most effective method for diver communication \cite{liu2022underwater}. However, when submerged underwater, the divers must constantly move all limbs to maintain body postures against the water, making it physically demanding and mentally exhausting to spare extra attention to hand gestures or tool operations. The water greatly limited divers' sense of the environment while restricting regular verbal communications or facial expressions, making it urgently necessary to develop novel solutions for intention recognition underwater for effective human-robot interactions \cite{Birk2022}.

    \begin{figure}[t]
        \centering
        \includegraphics[width=1\columnwidth]{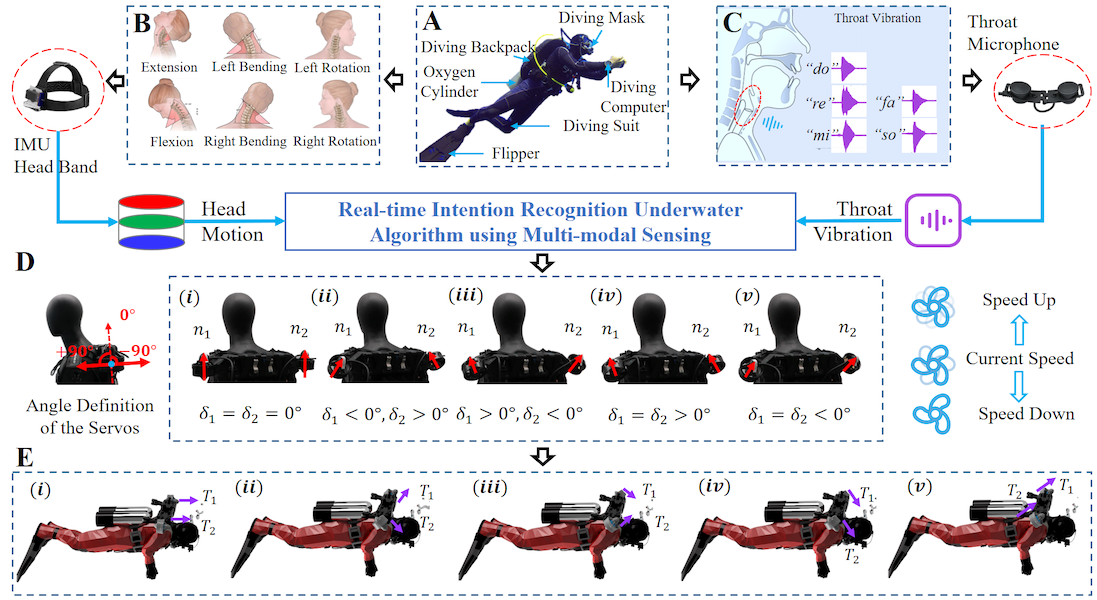}
        \caption{\textbf{Summary of the underwater intention recognition method using wearable IMU and throat microphone.} 
        (A) Diver with standard diving equipment, including a diving backpack, diving mask, diving computer, diving suit, flipper, oxygen cylinder, etc. 
        (B) The IMU headband can collect the motion information of the six types of head motion, including extension/flexion, bending left/right, and rotating left/right. 
        (C) The throat microphone can acquire the throat vibration, including the first five musical scales (``do'', ``re'', ``mi'', ``fa'', and ``so''). 
        (D) Based on the servo angle definition of the superlimb,  we define five types of motion modes. $\delta_1$ and $\delta_2$ are the angles of the left and right thrusters. $n_1, n_2$ and $T_1, T_2$ are the rotational speed and the thrust force of the left and right thruster, which can be controlled continuously via PWM.
        (E) The five motion modes mapping from the classification token with head motion or/and throat vibration.}
        \label{fig:PaperOverview}
    \end{figure}

    One way to drive novel designs for aquatic systems is by drawing inspiration from on-land systems to underwater applications, where the integration of head motion and throat vibration seems viable. For example, recent work by Yang \cite{yang2023mixed} demonstrated an artificial throat to check vocal vibrations to recognize everyday words vaguely spoken by a patient after a laryngectomy. Wang \cite{Wang2021} proposed a method through detecting eye motions and throat vibrations to interpret the intention of patients with amyotrophic lateral sclerosis (ALS). Severin \cite{severin2021head} developed a system using inertial sensors to detect head movement for intention recognition. Machangpa \cite{machangpa2018head} designed a wheelchair controlled by head gestures for quadriplegic patients. Although the divers' limbs are busy maintaining body postures and the mouth is filled with the breathing mouthpiece, we can leverage such limitations to use the head and throat to express intentions for controlling wearable robots underwater, such as an underwater superlimb \cite{2023ReconfigurableDesign}. 
    
    In this paper, we propose a novel solution for underwater intention recognition by simultaneously detecting the diver's head motion and throat vibration, as shown in Fig. \ref{fig:PaperOverview}, to enable multi-modal human-robot interactions with an underwater supernumerary robotic limb designed for providing propulsion assistance. The design features customization of the headband with a waterproof IMU sensor mounted on the top and a throat microphone on the neck for hands-free interaction. The system determines the diver's intention by sensing the diver's head motion through the IMU sensor, or confirms the control commands by detecting the diver's vocal vibration through the throat microphone using learning algorithms. By designing mapping commands to the underwater superlimb, the system recognizes the diver's intention for posture control underwater, aiming at reducing the diver's physical load and mental fatigue for nature interactions without using hands. The contributions of this study are as the following: 
    \begin{itemize}
        \item Proposed a novel design for underwater intention recognition by sensing the diver's head motion and throat vibration in a compact form factor for diving scenarios. 
        \item Developed a multi-modal, real-time classification algorithm based on five musical scales and six head motion types for intention recognition underwater. 
        \item Verified the feasibility of the proposed method for controlling an underwater superlimb prototype with continuous motion commands for underwater propulsion assistance.
    \end{itemize}
    
    The rest of this paper is organized as the following. Section \ref{sec:Method} presents the diver intention recognition method of the wearable sensing device, including the engineering design and classification algorithm for head motion and throat vibration. Section \ref{sec:Results} reports the experiment results using head motion, throat vibration, and combined modalities for superlimb control. Section \ref{sec:Discussion} discusses the experiment results and implications. The conclusion, limitations, and future work are in the final section.

\section{Method}
\label{sec:Method}
\subsection{Engineering Design}
    
\begin{figure}[t]
        \centering
        \includegraphics[width=1\columnwidth]{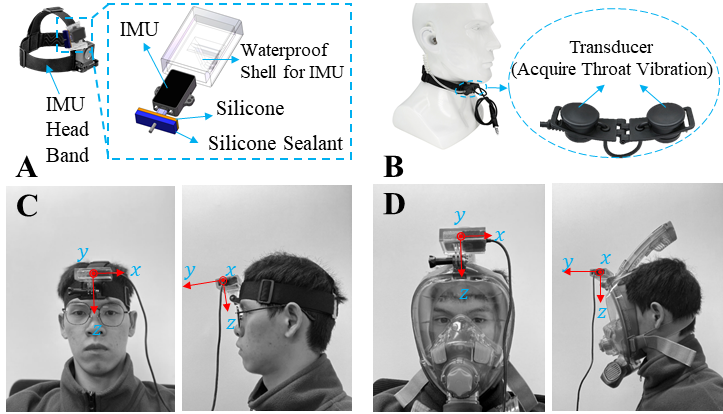}
        \caption{\textbf{Engineering design of the wearable sensing devices to collect the head motion and throat vibration data underwater.} 
        (A) IMU headband with a waterproof shell integrated with a 9-axis IMU (3DX-GX3-25 from Parker Hannifin), sealed by a silicone layer and a silicone sealant layer. 
        (B) Throat microphone (Z033 from WADSN Corp.) smeared with a polyurethane waterproofing spray (from SKSHU). 
        (C) A test user is wearing the waterproof IMU sensor with the headband (on land). 
        (D) The IMU sensor mounted on a full-face diving mask (for underwater).}
        \label{fig:SensingSystem}
    \end{figure}

    We designed a multi-modal sensing system as shown in Fig. \ref{fig:SensingSystem} for underwater intention detection. The IMU sensor can be fixed on the head using the headband, as shown in Figs. \ref{fig:SensingSystem}A\&C, picking up Euler angles and accelerations of head motion at up to 500Hz in 16 bits. The throat microphone is wearable on the neck to detect throat vibration at 16k or 60kHz in 16 bits, as shown in Fig. \ref{fig:SensingSystem}B. To protect the IMU from water eroding, we sealed the waterproof shell with silicone and sealant (Epoxy sealant for seawater from ROVMAKER). We modified the mask design (M8038 from SMACO), as shown in Fig. \ref{fig:SensingSystem}D, to adjust the IMU sensor's angle by turning the knob of the joint on top. The design of the IMU headband is compatible with the full-face diving mask for SCUBA divers, which connects the oxygen tank with a regulator for SCUBA diving.

\subsection{Detecting Head Motions and Throat Vibrations}

\begin{figure}[t]
        \centering
        \includegraphics[width=1\columnwidth]{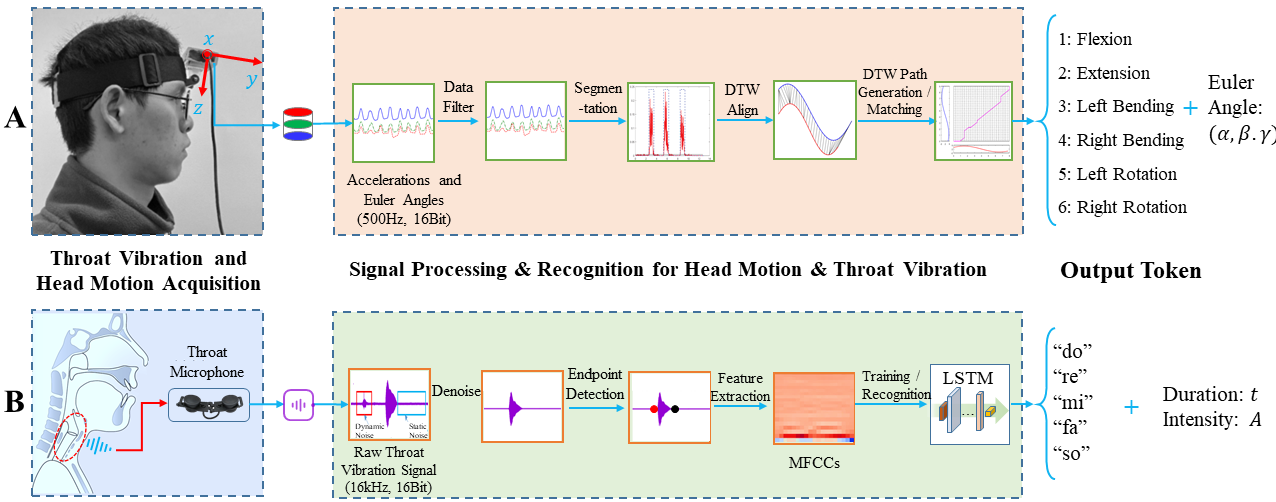}
        \caption{\textbf{Underwater interaction method based on the wearable sensing devices integrated with throat microphone and IMU.} 
        (A) The IMU can sense the head motion information, including acceleration and Euler angles. After the endpoint detection method based on adaptive thresholds, segments would be matching using DTW algorithm to distinguish the head motion types based on the head motion templates. 
        (B) The throat microphone can acquire the vibration of the throat. After noise reduction, the significant fragments of the raw signal are extracted through the endpoint detection method. Mel-filter bank analysis transforms these fragments into Mel-frequency cepstral coefficients (MFCCs). After LSTM processing and classification recognition, the command index can be mapped to a user-defined motion mode sent to control the superlimb.}
        \label{fig:HeadThroat}
    \end{figure}

\begin{figure}[t]
        \centering
        \includegraphics[width=1\columnwidth]{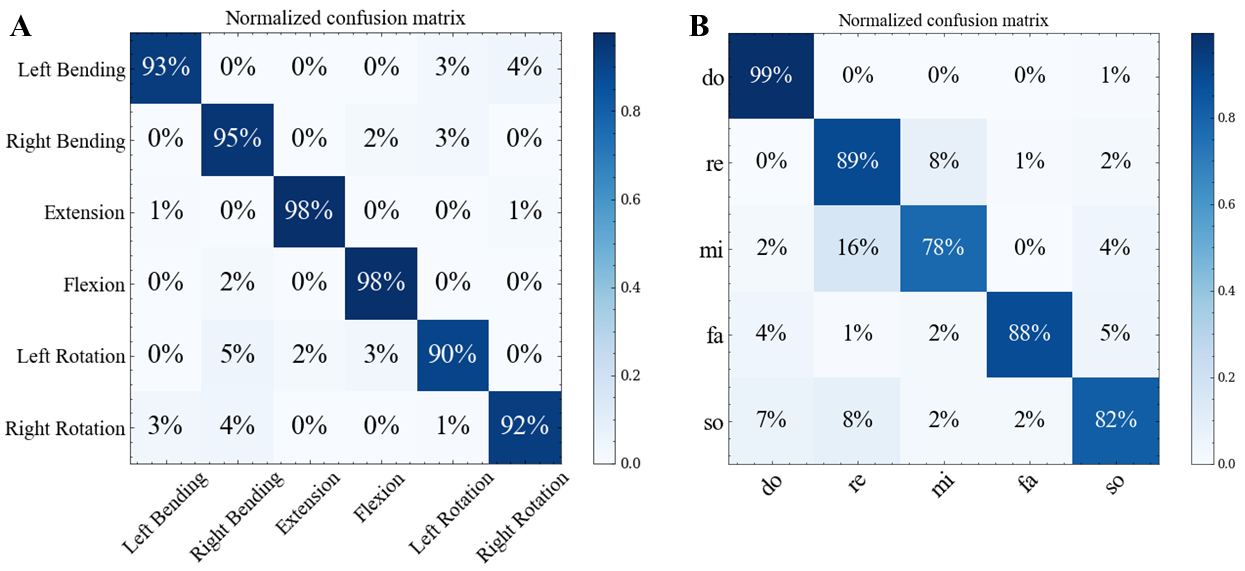}
        \caption{\textbf{Accuracy of recognition-related performance in the classification experiments.} 
        (A) Confusion matrix of head motion classification with IMU.
        (B) Confusion matrix of throat vibration classification with a throat microphone. 
        }
        \label{fig:ConfusionMatrix}
    \end{figure}

    We used different methods to process the two time-series data, as shown in Fig. \ref{fig:HeadThroat}. We adopted the Dynamic time warping (DTW) algorithm to distinguish six types of head motion (Bending left/right, Rotating left/right, Extension/Flexion), commonly used to process IMU data \cite{ krzeszowski2014DTW}. The raw data from the IMU (Accelerations and Euler Angles) was smoothed by a low-pass filter. Then, a self-adapting threshold segmentation method extracted the segment with the practical meaning of instructions. The DTW algorithms maximize the difference between different head motion types and minimize the distance between those of the same kind \cite{Senin2008DTW}. Since the Adaptive DTW barycenter averaging (ADBA) algorithm can average the motion data sequences in time and space, this time-series averaging template method has a higher recognition accuracy than a randomly averaged method. It was used to generate the data sequence template for the six head motion types \cite{2022inertial}. 1,436 head motion samples were collected from two male and two female participants (Bending left/right: 280/288, Extension/Flexion: 270/266, Rotating left/right: 296/306). Half of the dataset was used to generate the head motion template, and the rest was used for testing. Results show that the head motion recognition accuracy is measured at an average of $94\%$, as shown in Fig. \ref{fig:ConfusionMatrix}A.
    
    However, the result was unsatisfactory when we applied the same method for throat vibration data. Instead, we adopted the Mel-frequency cepstral coefficients (MFCCs) to extract features in speech recognition \cite{tiwari2010mfcc}. Alternatively, we can also use Long Short Term Memory (LSTM) as a candidate algorithm for acoustic modeling of speech \cite{Greff2017}. We collected throat vibration signals using a throat microphone and the Mel filter banks to transform the audio signal to MFCCs. After pre-processing the raw data, we obtained a 20 $\times$ 20 matrix by cutting off the MFCCs matrix or padding the zeros matrix into the time dimension of the MFCCs matrix, which describes the response of the human auditory system for the specific audio signal. Then, we fused the MFCCs matrix as the input of LSTM to get a classification result for the throat vibration. Ten participants (seven males, and three females) were invited to collect the throat vibration signal for data acquisition. They were asked to phonate musical scales with the throat microphone shown in Fig. \ref{fig:SensingSystem}B. We collected a dataset of 3,253 musical scale audio segments in WAV format containing 647 ``do'', 660 ``re'', 594 ``mi'', 668 ``fa'', and 684 ``so'', which were then split with 70\% for training and 30\% for testing. The model's average accuracy for testing is about $86\%$. The confusion matrix of the classification results is shown in Fig. \ref{fig:ConfusionMatrix}B.

\section{Results}
\label{sec:Results}
\subsection{Intention Recognition via Head Motion}

    We divided the head motions into two groups to control the speeds and angles of the two thrusters, respectively, as shown in Table \ref{tab:mapping_exp_1}. 
    \begin{table}[hbtp]
        \centering
        \caption{Mapping between the head motions with superlimb control.}
        \begin{threeparttable}
        \begin{tabular}{lcc}
        \hline
        Head Motion    & Euler Angle & Thruster Speeds$^{4}$ \\ 
        \hline
        Flexion        & $\alpha$    & $(-K_1\ast\alpha, -K_1\ast\alpha)$  \\
        Extension      & $\alpha$    & $(K_1\ast\alpha, K_1\ast\alpha)$ \\
        \hline
        Head Motion  & Euler Angle & Angle of Servos$^{1}$ \\
        \hline
        Left Bending   & $\beta$     & $(K_2 \ast\beta,K_2 \ast\beta)$ \\
        Right Bending  & $\beta$     & $(-K_2 \ast\beta,-K_2 \ast\beta)$  \\
        Left Rotation  & $\gamma$    & $(-K_3 \ast \gamma,K_3 \ast \gamma)$ \\
        Right Rotation & $\gamma$    & $(K_3 \ast \gamma,-K_3 \ast\gamma)$ \\
       
        \hline
        \label{tab:mapping_exp_1}
    
        \end{tabular}
        \begin{tablenotes}
        \footnotesize
        \item[1] Angle of Servos are those of the \textit{left} and \textit{right} ones. 
        \item[2] $K_1$ is the degree coefficient for $\alpha$.
        \item[3] $K_2$ is the degree coefficient for $\beta$.
        \item[4] Thruster Speeds are those of the \textit{left} and \textit{right} ones.
        \item[5] $K_3$ is the coefficient converting $\gamma$ to the thrusters speed.
        \end{tablenotes}
       \end{threeparttable}
    \end{table}
    
\begin{figure}[thpb]
        \centering
        \includegraphics[width=1\columnwidth]{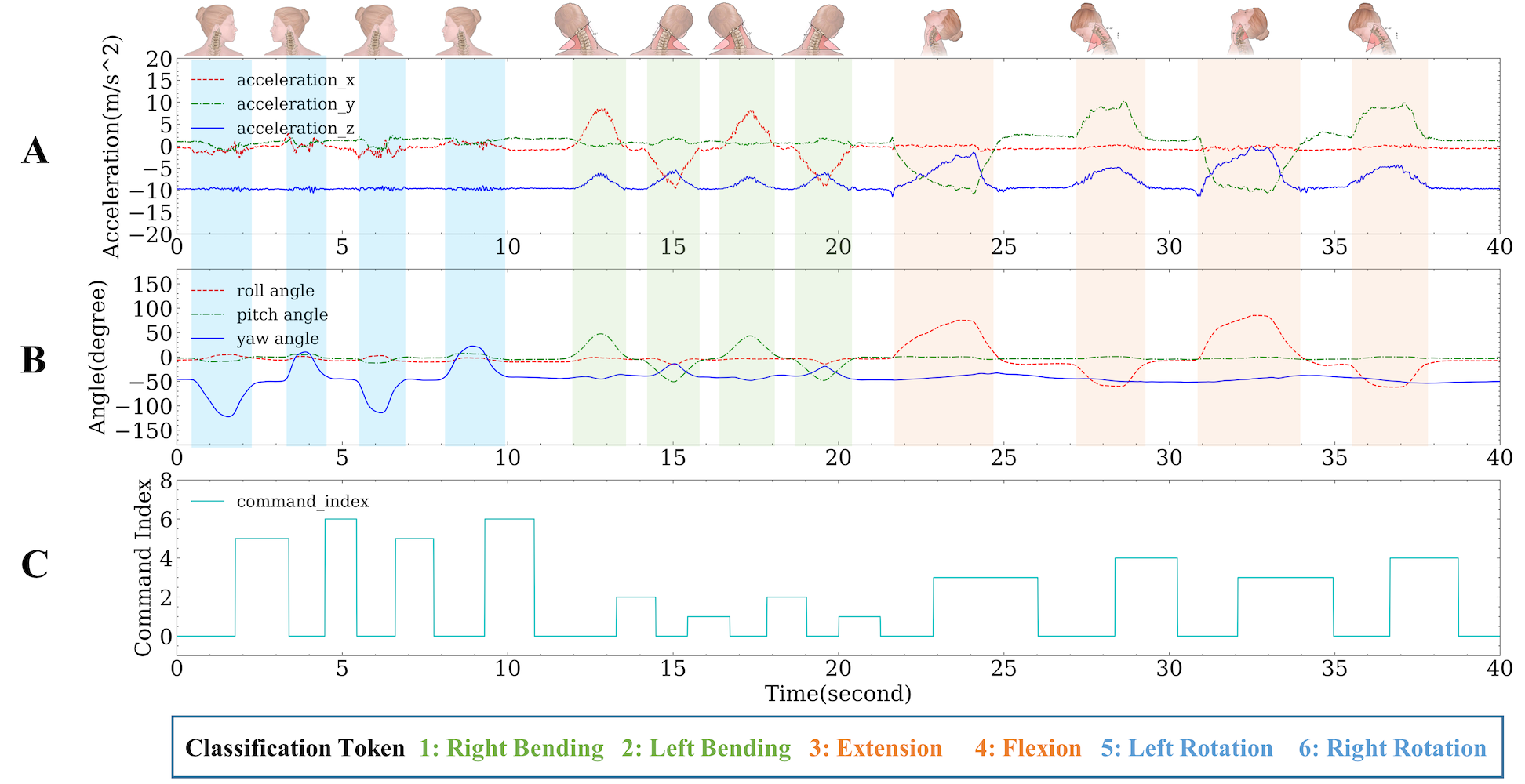}
        \caption{\textbf{Control of the superlimb using head motion.} 
        (A) Accelerations along the $x-y-z$ axis of the head motion from the IMU during the control process of the superlimb. 
        (B) Euler angles along the $x-y-z$ axis of the head motion from the IMU during the control process of the superlimb. 
        (C) According to the motion information of the data sequence by the head motion recognition algorithm, the head motion types were classified and the mapped motion control command was sent to the control unit of the superlimb.}
        \label{fig:Result-Head-A}
    \end{figure}

    Fig. \ref{fig:Result-Head-A} demonstrates the human-robot interactions experiments. The time series of accelerations and Euler angles along the $x/y/z$ axis recorded by IMU are shown in Figs. \ref{fig:Result-Head-A}A\&B. The corresponding control command sequence is shown in Fig. \ref{fig:Result-Head-A}C, where the mapping of head motions to control command index is (Bending right/left, Extension/Flexion, Rotating left/right) $\mapsto$ command index: $(1, 2, 3, 4, 5, 6)$. We executed four actions in each of the three DoFs of the rotation. The Euler angles range smoothly within [45°,50°] for Flexion, [70°,80°] for Extension, [35°,45°] for Bending left/right, and [65°,75°] for Rotating left/right, respectively. The system recognized all 12 head motions correctly, and the corresponding control commands were sent to the superlimb afterward. 
    
\begin{figure}[thpb]
        \centering
        \includegraphics[width=1\columnwidth]{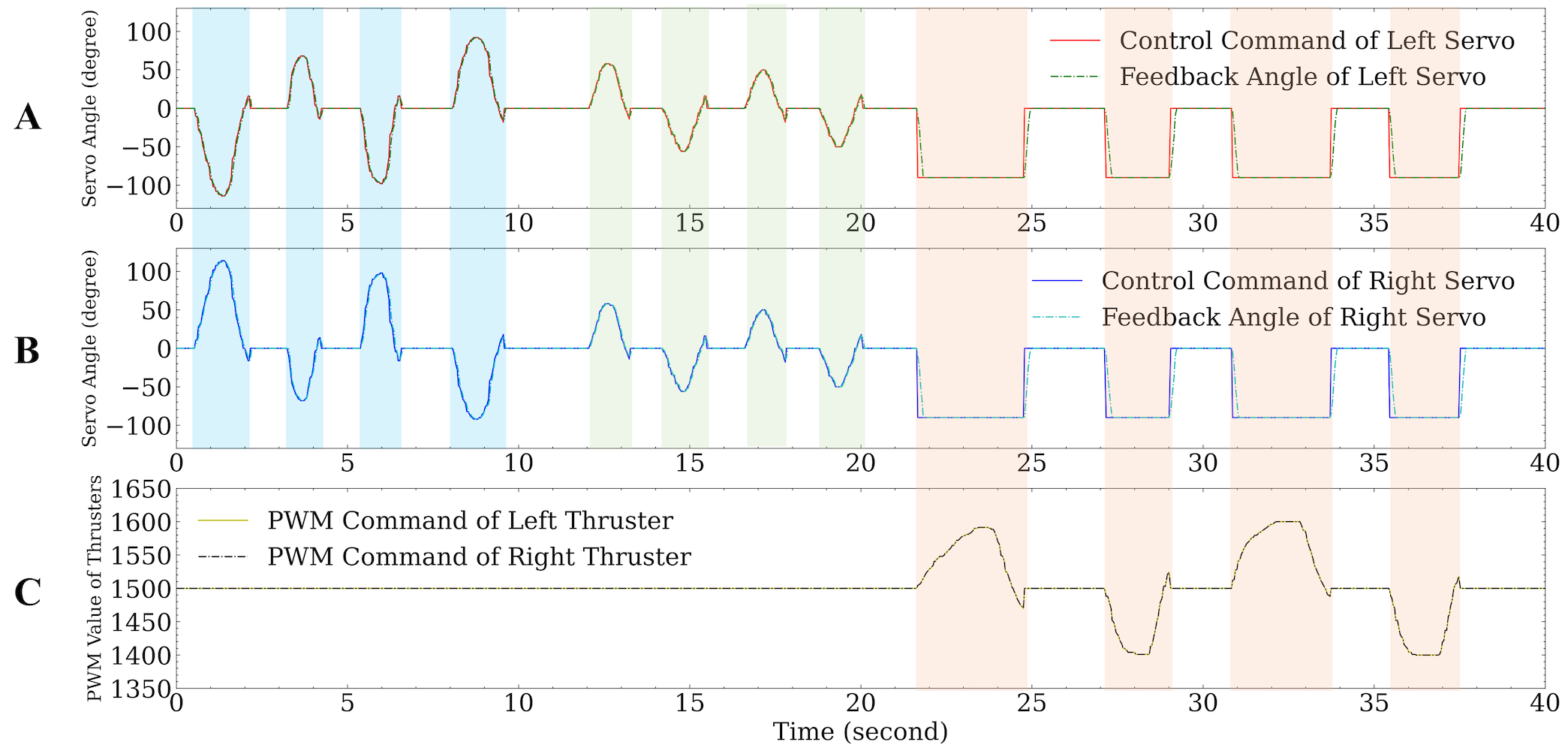}
        \caption{\textbf{Experimental results of superlimb control with head motion recognition experiment: comparing theoretical output and actual feedback of the superlimb .} 
        (A) The Control command and the actual feedback of the left servo. 
        (B) The Control command and the actual feedback of the right servo. 
        (C) The PWM command is sent to the left and right thruster based on the classification token output from the head motion recognition algorithm.}
        \label{fig:Result-Head-B}
    \end{figure}

    Fig. \ref{fig:Result-Head-B} compares the control command and actual feedback of the servos and thrusters from the superlimb, indicating that the pipeline can detect human intentions and achieve robot control continuously with low latency (less than one second). However, to achieve precise control of the thrusters through the Euler angles of the head motion, an operator would need training and practice to obtain muscle memories of finer-grained mapping between head motions and robot control. Such activity involves humans in the loop as a human-robot system though the robot system alone is an open control loop.
    
\subsection{Intention Recognition via Throat Vibration}

    We designed a mapping between musical scales and both thrusters' angle and speed for the throat vibration signal in Table \ref{tab:mapping_exp_2}. Three types of musical scales and lengths distinguish six kinds of control commands. Meanwhile, the amplitude $A$ of the musical scale signal (within 64 $ms$) is detected in real-time continuously and is used as a coefficient in each of the six commands. For example, a short type of ``do'' is mapped to rotating both thrusters to positive angles (show in Fig .\ref{fig:PaperOverview} E($v$))  determined by its amplitude $A_1$.

    \begin{table}[hbtp]
        \centering
        \caption{mapping superlimb control with throat vibration recognition experiment}
         \begin{threeparttable}
        \begin{tabular}{lcccc}
        \hline
        Musical Scale & Duration  &  Amplitude &  Angle of Servo$^1$\\
        \hline
        do  & $t<500ms$    & $A_1$   & $(A_1\ast K_4,\enspace\,\, A_1\ast K_4)$    \\
        do  & $t>500ms$    & $A_2$   & $(-A_2\ast K_4,-A_2\ast K_4)$       \\
        re  & $t<500ms$    & $A_3$    & $(-A_3\ast K_4 ,\enspace\,\, A_3\ast K_4)$        \\
        re  & $t>500ms$    & $A_4$     & $(A_4\ast K_4,-A_4\ast K_4)$      \\
        \hline
        Musical Scale    & Duration  &Amplitude & Thruster Speeds$^3$    \\ 
        \hline
        mi  & $t<500ms$    & $A_5$   & $(A_5\ast K_5,\enspace\,\, A_5\ast K_5)$ \\
        mi  & $t>500ms$    & $A_6$      & $(-A_6\ast K_5,\enspace\,\, -A_6\ast K_5)$      \\
        \hline
        \label{tab:mapping_exp_2}
        \end{tabular}
         \begin{tablenotes}
        \item [1] Angle of Servos are those of the \textit{left} and \textit{right} ones. 
        \item [2] $K_4$ is the degree coefficient for throat vibration amplitude.
        \item [3] Thruster Speeds are those of the \textit{left} and \textit{right} ones. 
        \item [4] $K_5$ is the coefficient converting throat vibration amplitude to thruster speed.
     \end{tablenotes}
        \end{threeparttable}
    \end{table}

\begin{figure}[thpb]
        \centering
        \includegraphics[width=1\columnwidth]{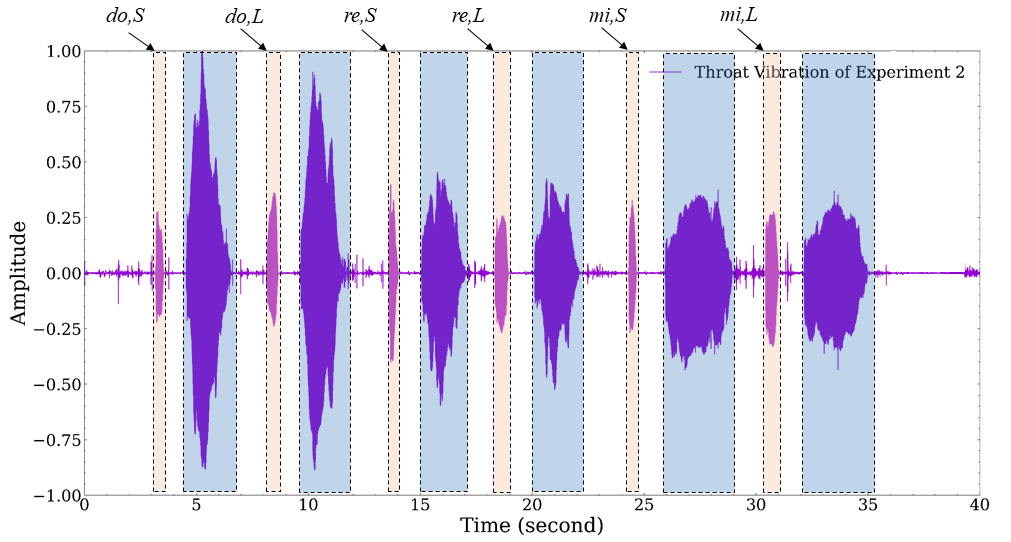}
        \caption{\textbf{Waveform of throat vibration acquired by throat microphone during superlimb control with throat vibration recognition experiment.} Marked in pink shaded areas are the different types of throat vibration, and marked in blue shaded areas are the throat vibrations used to continuously control the servo angle and thruster speed of the superlimb according to the amplitude of the vibration signals.}
        \label{fig:Result-Throat-A}
    \end{figure}

\begin{figure}[thpb]
        \centering
        \includegraphics[width=1\columnwidth]{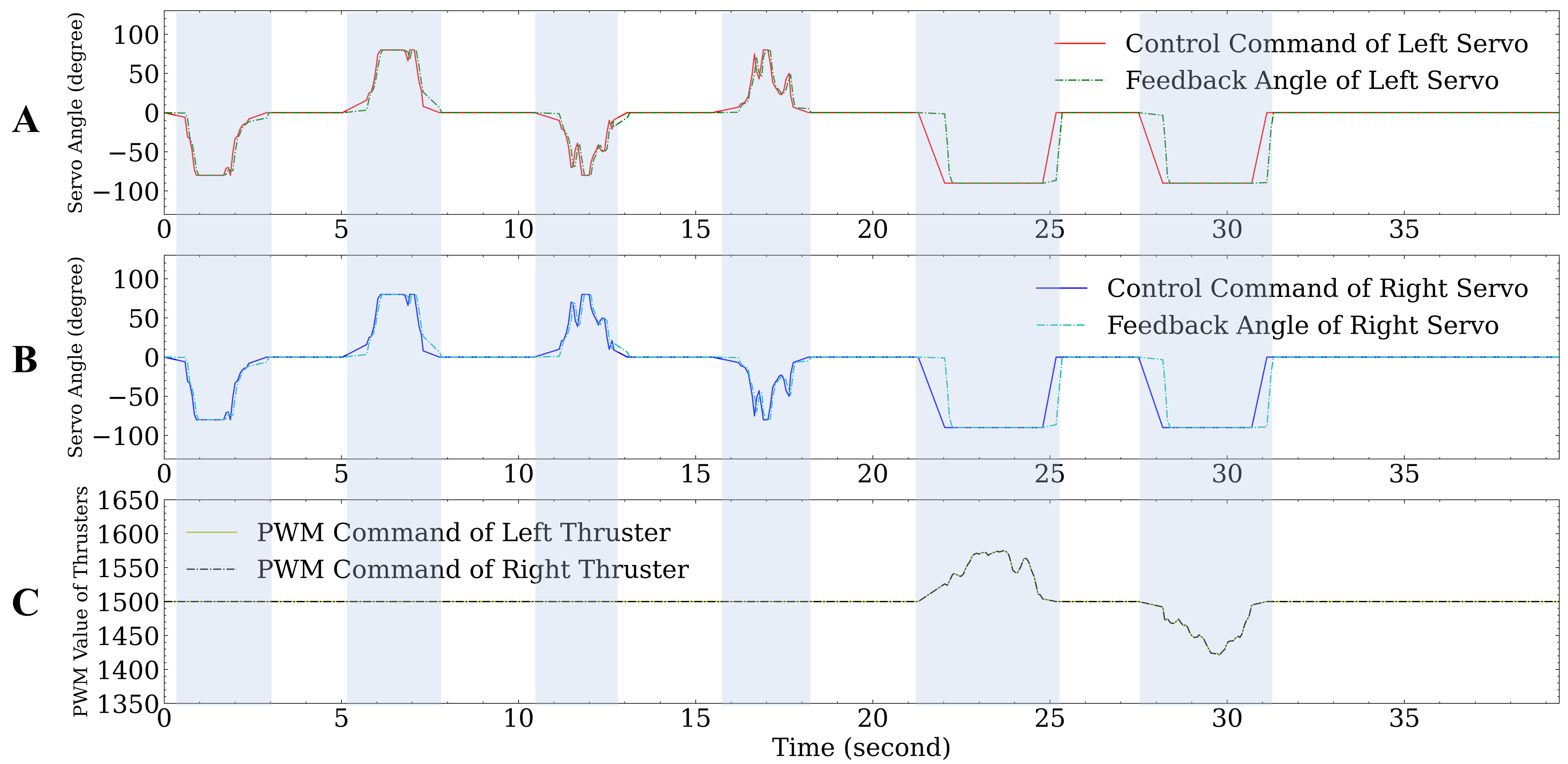}
        \caption{\textbf{Experimental results of superlimb control with Throat vibration recognition experiment: comparing theoretical output and actual feedback of the superlimb.} 
        (A) The Control command and the actual feedback of the left servo. 
        (B) The Control command and the actual feedback of the right servo. 
        (C) The PWM command is sent to the left and right thruster based on the classification token output from the throat vibration recognition algorithm.}
        \label{fig:Result-Throat-B}
    \end{figure}

    We demonstrate the human-robot interaction through throat vibration. Fig. \ref{fig:Result-Throat-A} shows the raw signal (solid purple line) and the recognized musical scales. Every intention consists of two sequential waveform segments, with the first indicating the type of command (pink shaded areas) and the second for the amplitude of action (blue shaded areas). Although one could still express control intentions with the throat vibration, the user must be trained in vocal control for the system to recognize the intention effectively. Fig. \ref{fig:Result-Throat-B} shows the theoretical and actual feedback of the servo angles and the control command of PWM (Range from $[1100,1900]$) sent to the control module of the thrusters. The blue shaded area is the action execution of the superlimb based on the corresponding motion mode shown in Fig. \ref{fig:Result-Throat-A}.
    
\subsection{Multi-modal Intention Recognition and Interactions}
    
    In this experiment, we test the feasibility of using head motion and throat vibration simultaneously as a multi-modal mechanism for controlling the underwater superlimb based on intention recognition. Table \ref{tab:mapping_exp_3} defines the action vectors to describe the diver's robot control intentions mapped to the throat vibration and head motion. 
    \begin{table}[hbtp]
        \centering
        \caption{Mapping of multi-modal intention recognition and interactions experiment.}
        \begin{threeparttable}
        \begin{tabular}{lll}
        \hline
        Action vector              & Left thruster (rpm)  & Right thruster (rpm)      \\
        \hline
        (do,short,null)            & Accelerate $k$              & null              \\
        (do,long,null)             & Accelerate -$k$              & null                \\
        (re,short,null)            & null                             & Accelerate $k$   \\
        (re,long,null)             & null                             & Accelerate $-k$  \\
        (mi,short,null)            & Stop                             & Stop                \\
        (mi,long,null)            & Stop                           & Stop                \\
        (fa,short,null)            &  Accelerate $k$              & Accelerate $k$   \\
        (fa,long,null)             & Accelerate $-k$              & Accelerate $-k$  \\
        (so,short,null)            & Switch control mode              & Switch control mode \\
        (so,long,null)            & Switch control mode              & Switch control mode \\
        \hline
        Action vector              & Left servo (degree)  & Right servo (degree)         \\
        \hline
        (null,null,left rotation)  & $-90 $                        & null                \\
        (null,null,right rotation) & $\enspace\,\, 90 $            & null                \\
        (null,null,left bending)   & $\enspace\,\,90 $             & null                \\
        (null,null,right bending)  & $-90 $                        & null                \\
        (null,null,extension)      & $-90 $                         & $-90$         \\
        (null,null,flexion)        & $\enspace{\,\,}90 $         & $\enspace\,\,90$           \\
        \hline
        \label{tab:mapping_exp_3}
        \end{tabular}
        
        \begin{tablenotes}
        \item [1] $null$ means tester has no action or superlimb doesn't change its motion.
        \item [2] $k$ means the acceleration coefficient of the thrusters.
        \end{tablenotes}
        \end{threeparttable}
    \end{table}
    
\begin{figure}[thpb]
        \centering
        \includegraphics[width=1\columnwidth]{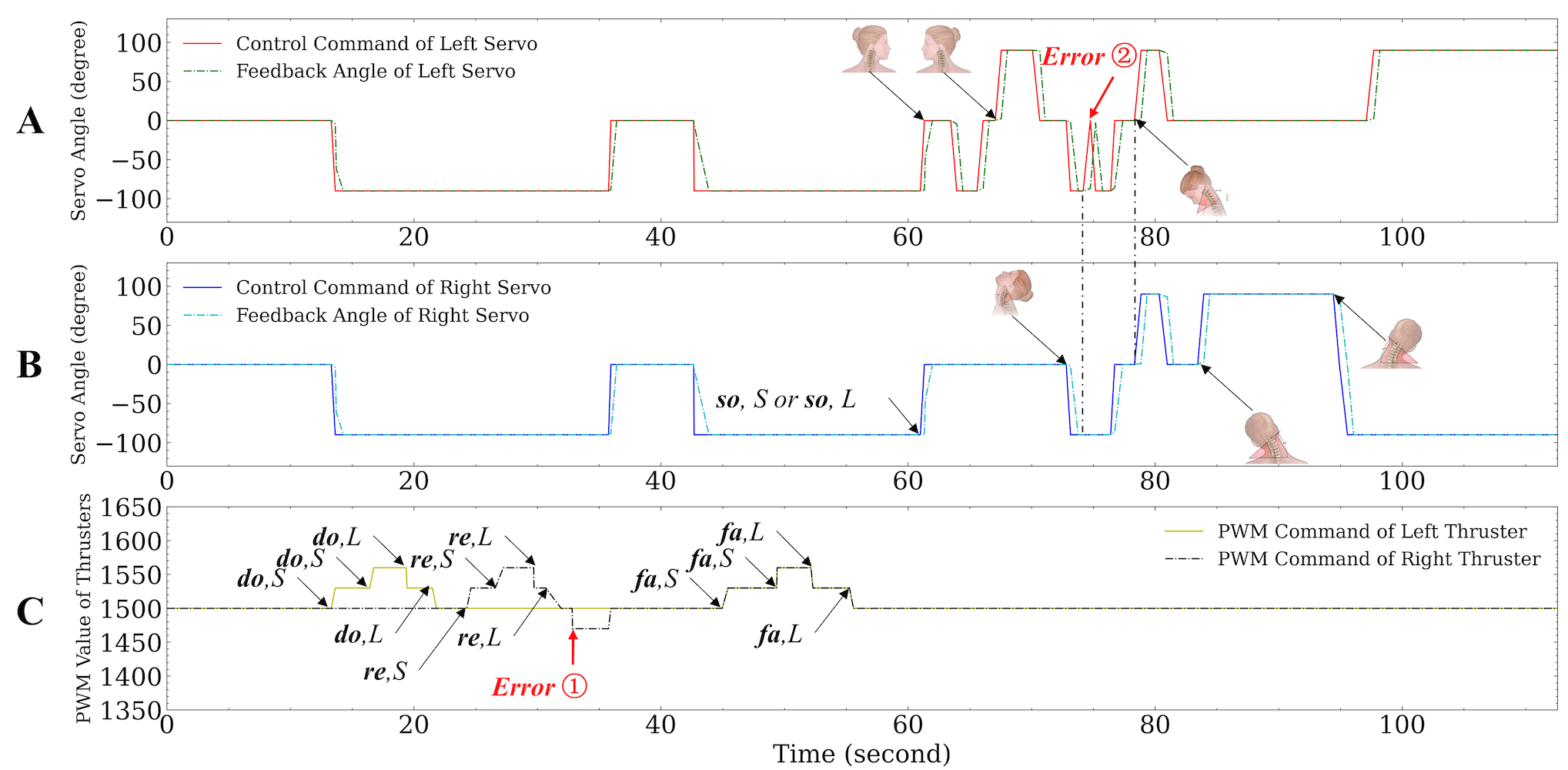}
        \caption{\textbf{Experimental results of multi-modal intention recognition and interactions experiment: comparing theoretical output and actual feedback of the superlimb using head motion and throat vibration.} 
        (A) The control command and the actual feedback of the left servo.
        (B) The control command and the actual feedback of the right servo. 
        (C) The PWM command sent to the left and right thrusters based on the token output from the intention recognition algorithm.}
        \label{fig:Result-Combine-Head-Throat}
    \end{figure}

    In this experiment, the musical scale ``so'' was defined as the mode switch from the control of servo angle to the rotational speed of the thruster. Fig. \ref{fig:Result-Combine-Head-Throat} compares theoretical output and actual measurement of the superlimb. We made two observations in the multi-modal integration experiment. The noticeable latency of the system could increase to two seconds when controlling the servos using head motion because of the different frequencies between these two recognition methods. The misalignment in Fig. \ref{fig:Result-Combine-Head-Throat}C is caused by the system transforming $(re,long,null)$ to $(so,long,null)$. The other observation is when the extension was not correctly recognized in Fig. \ref{fig:Result-Combine-Head-Throat}, which was caused by an unexpected servo angle during the experiment. After all, the head motion recognition module has to lower the frequency of data collection and classification to meet the frequency of the throat vibration recognition module.
    
\section{Discussion}
\label{sec:Discussion}
\subsection{Towards Underwater Intention Recognition}
    
    This study presents the engineering design and experiment results of an underwater multi-modal interaction mechanism for intention recognition using head motions and throat vibrations. Although the reported system is still a lab prototype that requires further testing underwater, the simple design in a compact form factor makes the proposed solution promising for human-robot interactions underwater. For on-land scenarios, the head motion or the throat vibration has been demonstrated effective in intention recognition for different applications with no need for integration. In this study, due to the need for aquatic interaction, we propose to combine these two modalities for intention recognition underwater. The high classification performance reported in this study aligns with the literature. Our results further demonstrated that when combining these two modalities, they formulate an intuitive mechanism for intention expression that is effective in intention recognition through learning algorithms, which could be a practical solution for controlling an underwater superlimb robot.

    This study tested only three pairs of head motion and five musical scales for intention expression and recognition. However, one can quickly expand the vocabulary by extending the head motion to the total five degree-of-freedoms of the head to include the two translational motions. Some users may need further training before being able to do so fluently. On the other hand, one can also expand the musical scales to a broader range or develop the system to recognize a sequence of them. For example, it is easier for people to remember a piece of tune rather than the specific musical scale for differentiating different meanings. The artificial throat \cite{yang2023mixed} provides an excellent inspiration to expand this work towards a more natural expression of intentions for the system to recognize effectively, which we intend to further explore by using sensors of more compact sizes \cite{2023GUOChuanfei}.

\subsection{Learning Intentions via Head Motion \& Throat Vibration}

    In this study, we proposed a multi-modal learning framework for integrating head motion and throat vibration in intention recognition. For on-land scenarios with a clear voice, one could directly use conventional methods to differentiate the volume, pitch, and tune from voice signals, a mature technology already used in commercial products. However, the experiments in this study specifically chose a learning approach as the signal detected was through the throat, which had more noise than those collected from the mouth. On the other hand, when submerged underwater, the noise signal from the water would further reduce the quality of the sound signals, making it a challenging task to classify different vocal signals clearly. However, our experiment results show that the learning algorithms effectively classified the different musical notes hymned by the test user. We intend to further test the proposed system by collecting training data from underwater to refine the model for a more realistic scenario. 

    On the other hand, the experiment results of the combined modalities in controlling the superlimb robot were successful, real-time, and continuous (please refer to the supplementary materials for demonstration). Throughout the experiment, only the head movement and throat vibrations were used to support hands-free interaction with the underwater superlimb. Further testing in the aquatic environment is needed in future work to test the proposed system's performance thoroughly. On the other hand, one can easily extend the application of the proposed method to interact with other underwater robots, such as UAVs and robotic fishes, on-land robots, such as robotic manipulators, legged robots, aerial robots, or mobile robots, or common Internet-of-Things (IoTs) devices in domestic scenarios. Another application area is for people with vocal impairment, where slight modifications could make the system wearable for users with disabilities. 

\subsection{Human-Robot Interactions for an Underwater Superlimb}

    Humans are not biologically evolved for natural activities underwater, which influenced the priority of design considerations when developing diving gear. The current diving devices mainly provide life-support and swimming assistance in an aquatic environment while being wearable to the diver's body forms. With the limited space, the complexity of need, and the waterproof requirement, two significant challenges remain in introducing robotic intelligence to diving gear. One is the design problem for a wearable robot compatible with the existing diving gear while providing meaningful assistance underwater to reduce the physical load of the body limbs. In our previous work \cite{2023ReconfigurableDesign}, we developed a reconfigurable underwater jetpack to enable wearable propulsion with compatible connections to the current Buoyancy control device (BCD) system, aiming at sharing the burdens of manual posture control underwater so that the diver can spare their hands for tool operation. We named it an underwater superlimb due to its functionalities in providing supernumerary limb support for manual posture control during diving and swimming, similar to other superlimb designs for on-land scenarios \cite{kurek2017design}. 

    Intention recognition is the other problem that is yet to be solved when developing an intelligent wearable underwater robot, which is shared by the other superlimbs for on-land scenarios. The proposed solution, while being inspired by many recent works for on-land scenarios, is found to be practical for underwater human-robot interactions. With limited sensory feedback and limited eyesight underwater, divers usually need to turn their heads constantly for a better inspection of the surrounding environment, providing a more explicit cue for expressing intentions without using their hands. While the diver's mouth is usually filled with breathing tubes of the regulator, common language expression becomes a challenge. Even with a full-face mask, there are still difficulties in underwater signal or voice transmission. However, as demonstrated in this study, one can still leverage the throat vibration to express a wide range of vocal commands for intuitive and direct interaction.

\section{Conclusions}
\label{sec:FinalRemarks}

    In this study, we proposed a novel mechanism for underwater intention recognition using head motion and throat vibration. Experiment results showed that the system accurately classified different intention expressions coded through these two signals. The proposed multi-modal learning algorithm effectively recognized the intentions of the test user and controlled the underwater superlimb robot through various commons. 

    This study still needs to be improved in collecting data from the underwater environments for training, which we intend to conduct once the water pools are open for testing. The current system was tested on a breadboard, which needs further integration with the underwater superlimb's controller for an integrated system design. Further refinement of the command mapping between these two modalities and the robot control commands could be arranged. Nevertheless, the results of this study paved the foundation for future development in underwater intention recognition and underwater human-robot interactions with supernumerary support. 
    
\addtolength{\textheight}{-12cm}   







\bibliographystyle{IEEEtran}
\bibliography{references}  

\begin{thebibliography}{10}
\providecommand{\url}[1]{#1}
\csname url@samestyle\endcsname
\providecommand{\newblock}{\relax}
\providecommand{\bibinfo}[2]{#2}
\providecommand{\BIBentrySTDinterwordspacing}{\spaceskip=0pt\relax}
\providecommand{\BIBentryALTinterwordstretchfactor}{4}
\providecommand{\BIBentryALTinterwordspacing}{\spaceskip=\fontdimen2\font plus
\BIBentryALTinterwordstretchfactor\fontdimen3\font minus
  \fontdimen4\font\relax}
\providecommand{\BIBforeignlanguage}[2]{{%
\expandafter\ifx\csname l@#1\endcsname\relax
\typeout{** WARNING: IEEEtran.bst: No hyphenation pattern has been}%
\typeout{** loaded for the language `#1'. Using the pattern for}%
\typeout{** the default language instead.}%
\else
\language=\csname l@#1\endcsname
\fi
#2}}
\providecommand{\BIBdecl}{\relax}
\BIBdecl

\bibitem{straughan2012touched}
E.~R. Straughan, ``Touched by water: The body in scuba diving,'' \emph{Emotion,
  Space and Society}, vol.~5, no.~1, pp. 19--26, 2012.

\bibitem{xia2022wearable}
H.~Xia, M.~A. Khan, Z.~Li, and M.~Zhou, ``Wearable robots for human underwater
  movement ability enhancement: A survey,'' \emph{IEEE/CAA Journal of
  Automatica Sinica}, vol.~9, no.~6, pp. 967--977, 2022.

\bibitem{liu2022underwater}
T.~Liu, Y.~Zhu, K.~Wu, and F.~Yuan, ``Underwater accompanying robot based on
  ssdlite gesture recognition,'' \emph{Applied Sciences}, vol.~12, no.~18, p.
  9131, 2022.

\bibitem{Birk2022}
A.~Birk, ``A survey of underwater human-robot interaction (u-hri),''
  \emph{Current Robotics Reports}, vol.~3, no.~4, pp. 199--211, 2022.

\bibitem{yang2023mixed}
Q.~Yang, W.~Jin, Q.~Zhang, Y.~Wei, Z.~Guo, X.~Li, Y.~Yang, Q.~Luo, H.~Tian, and
  T.-L. Ren, ``Mixed-modality speech recognition and interaction using a
  wearable artificial throat,'' \emph{Nature Machine Intelligence}, vol.~5,
  no.~2, pp. 169--180, 2023.

\bibitem{Wang2021}
Z.~Wang, A.~Zhang, B.~Chen, J.~Wang, and S.~Gao, ``A piezoelectric artificial
  throat and computer vision based on smart assistive system,'' in \emph{2021
  IEEE International Conference on Flexible and Printable Sensors and Systems
  (FLEPS)}, 2021, pp. 1--4.

\bibitem{severin2021head}
I.-C. Severin, ``Head gesture-based on imu sensors: a performance comparison
  between the unimodal and multimodal approach,'' in \emph{2021 International
  Symposium on Signals, Circuits and Systems (ISSCS)}.\hskip 1em plus 0.5em
  minus 0.4em\relax IEEE, 2021, pp. 1--4.

\bibitem{machangpa2018head}
J.~W. Machangpa and T.~S. Chingtham, ``Head gesture controlled wheelchair for
  quadriplegic patients,'' \emph{Procedia computer science}, vol. 132, pp.
  342--351, 2018.

\bibitem{2023ReconfigurableDesign}
J.~Huo, J.~Wang, Y.~Guo, W.~Qiu, H.~Asada, F.~Wan, and C.~Song,
  ``Reconfigurable design and modeling of an underwater superlimb for diving
  assistance,'' \emph{Advanced Intelligent Systems (Under Review)}, 2023.

\bibitem{krzeszowski2014DTW}
T.~Krzeszowski, A.~Switonski, B.~Kwolek, H.~Josinski, and K.~Wojciechowski,
  ``Dtw-based gait recognition from recovered 3-d joint angles and inter-ankle
  distance,'' in \emph{Computer Vision and Graphics: International Conference,
  ICCVG 2014, Warsaw, Poland, September 15-17, 2014. Proceedings}.\hskip 1em
  plus 0.5em minus 0.4em\relax Springer, 2014, pp. 356--363.

\bibitem{Senin2008DTW}
P.~Senin, ``Dynamic time warping algorithm review,'' \emph{Information and
  Computer Science Department University of Hawaii at Manoa Honolulu, USA},
  vol. 855, no. 1-23, p.~40, 2008.

\bibitem{2022inertial}
S.~Xu and S.~Lee, ``An inertial sensing-based approach to swimming pose
  recognition and data analysis,'' \emph{Journal of Sensors}, vol. 2022, pp.
  1--12, 2022.

\bibitem{tiwari2010mfcc}
V.~Tiwari, ``Mfcc and its applications in speaker recognition,''
  \emph{International journal on emerging technologies}, vol.~1, no.~1, pp.
  19--22, 2010.

\bibitem{Greff2017}
K.~Greff, R.~K. Srivastava, J.~Koutník, B.~R. Steunebrink, and J.~Schmidhuber,
  ``Lstm: A search space odyssey,'' \emph{IEEE Transactions on Neural Networks
  and Learning Systems}, vol.~28, no.~10, pp. 2222--2232, 2017.

\bibitem{2023GUOChuanfei}
J.~Shi, Y.~Dai, Y.~Cheng, S.~Xie, G.~Li, Y.~Liu, J.~Wang, R.~Zhang, N.~Bai,
  M.~Cai \emph{et~al.}, ``Embedment of sensing elements for robust, highly
  sensitive, and cross-talk--free iontronic skins for robotics applications,''
  \emph{Science Advances}, vol.~9, no.~9, p. eadf8831, 2023.

\bibitem{kurek2017design}
D.~Kurek, ``Design and control of supernumerary robotic limbs for near-ground
  work,'' Ph.D. dissertation, Massachusetts Institute of Technology, 2017.

\end{thebibliography}

\vfill

\end{document}